\documentclass[letterpaper, 10pt, conference]{ieeeconf}  

\IEEEoverridecommandlockouts                              

\usepackage[style=ieee, doi=false, url=false, eprint=false, isbn=false, natbib=true, mincitenames=1, maxcitenames=1, maxbibnames=99]{biblatex}
\usepackage{amsmath}
\usepackage{amssymb}
\usepackage[hidelinks]{hyperref}
\usepackage{color}
\usepackage{graphicx}
\usepackage{subcaption}
\usepackage{multirow}
\usepackage{booktabs}

\DeclareMathOperator*{\argmax}{arg\,max}

\newcommand{\ph}[1]{{\textbf{#1}:}} 

\newcommand{\fadhil}[1]{{\color{black}#1}}

\title{\LARGE \bf
Capability-aware Task Allocation and Team Formation Analysis for Cooperative Exploration of Complex Environments
}

\author{Muhammad Fadhil Ginting$^{1,2}$, Kyohei Otsu$^{2}$, Mykel J. Kochenderfer$^{1}$, and Ali-akbar Agha-mohammadi$^{2}$
\thanks{*This work was supported by the Jet Propulsion Laboratory, California Institute of Technology, under a contract with the National Aeronautics and Space Administration.}
\thanks{$^{1}$Department of Aeronautics \& Astronautics, Stanford University, Stanford, CA, USA
        {\tt\small {\{ginting, mykel\}}@stanford.edu}}%
\thanks{$^{2}$NASA Jet Propulsion Laboratory, California Institute of Technology, Pasadena, CA, USA
        {\tt\small {\{otsu, aliagha\}}@jpl.nasa.gov}}%
\thanks{\copyright 2022 California Institute of Technology. All rights reserved.}
}

\begin{document}

\maketitle

\begin{abstract}
To achieve autonomy in complex real-world exploration missions, we consider deployment strategies for a team of robots with heterogeneous autonomy capabilities. 
In this work, we formulate a multi-robot exploration mission and compute an operation policy to maintain robot team productivity and maximize mission rewards. 
The environment description, robot capability, and mission outcome are modeled as a Markov decision process (MDP). 
We also include constraints in real-world operation, such as sensor failures, limited communication coverage, and mobility-stressing elements. 
Then, we study the proposed operation model on a real-world scenario in the context of the DARPA Subterranean (SubT) Challenge. 
The computed deployment policy is also compared against the human-based operation strategy in the final competition of the SubT Challenge. 
Finally, using the proposed model, we discuss the design trade-off on building a multi-robot team with heterogeneous capabilities.

\end{abstract}

\section{Introduction}

Autonomous multi-robot systems have the potential to enable complex and ambitious missions that were not possible before. 
Prominent examples include NASA JPL's Mars 2020 rover and helicopter mission \cite{mars2020web} and autonomous multi-robot exploration in the DARPA Subterranean Challenge \cite{subt_webpage}.
This recent development opens possibilities for more elaborate missions that require exploration of large area and various challenging terrains beyond the capability of a single robot~\cite{touma2020mars}. 

While multi-robot systems provide resiliency and enable more complex missions, designing and deploying a robot team requires a significant study of the robots' capabilities and mission specifications. 
Deploying a team of multiple robots with different capabilities increases the complexity of the mission. 
The nature of exploration tasks requires various autonomous technologies as described in \cite{agha2021nebula}.
In this work, we aim to find an optimal robot team configuration and deployment strategy for a given exploration mission.


Our work focuses on multi-robot mission planning and task allocation, which is located at a high level in the autonomy stack. 
This module has to cope with various types of uncertainties in the environment and the robot's performance. 
The uncertainties come from new and updated information as the robot team explores the environment.  
Robot performance can be unpredictable due to mobility and sensory failures in diverse and challenging real-world settings. 
Moreover, communication is typically limited in a large-scale and complex environment and poses a coordination challenge between the robots. 

\begin{figure}[t]
    \centering
    \includegraphics[width=0.49\textwidth]{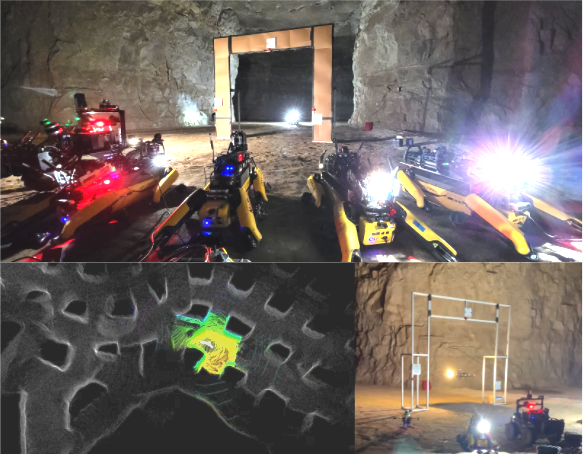}
    \caption{Robot team deployment underground in a Kentucky limestone mine. Robots with different mobility capabilities (i.e., wheeled, legged, aerial) are sent to the mine to collaboratively map the environment and find anomalies.}
    \label{fig:robot_team_in_mine}
\end{figure}

In this work, we develop an approach to model heterogeneous multi-robot exploration in a real-world scenario and compute an efficient deployment strategy.
We construct the model as a Markov decision process (MDP) in a combined space of environment description, robot capabilities, and mission outcome. 
The environment is modeled as a topological graph that has information on the map connectivity that takes robot mobility capabilities into account, 
we incorporate various realistic constraints such as communications and sensor failures to model real-world behaviors.
We perform case studies based on simulations and our real-world robot deployment experiences in the DARPA Subterranean (SubT) Challenge. 

Moreover, we show how our operation model can be used to determine the best robot team configuration and analyze the impact of different robot capabilities for a given mission.


Our contributions are summarized as follows:
\begin{itemize}
    \item We develop a novel problem formulation of unknown-space exploration by a heterogeneous robot team by modeling environments, robotic capabilities, and mission outcome as a MDP (\autoref{sec:problemsetup}).
    \item We demonstrate that our model can compute multi-robot operation policies in a real-world scenario and compare it against the human-based decisions in the live operation (\autoref{sec:simulation}).
    \item We analyze and discuss the design trade-off on formulating multi-robot teams with different mobility, perception, communication, and autonomy capabilities (\autoref{sec:exp-capability-analysis}).
\end{itemize}


\section{Related Works}
\label{sec:relworks}
\ph{Multi-robot coordination} Multi-robot task allocation is considered in different domains including multi-robot coverage planning~\cite{schneider1998territorial}, formation~\cite{alonso2017multi}, and collaborative manipulation~\cite{culbertson2018decentralized}. Some of the key aspects in designing and deploying multi-robot systems are sensor coverage, robot heterogeneity, communication, and task allocation. In this work we model these aspects to plan an optimal strategy for the robot team.
For the comprehensive review on the multi-robot approaches, we refer the reader to~\cite{parker2016multirobot}.

\ph{Multi-robot foraging and coverage}
Our work is mainly related to the multi-robot applications in foraging and coverage in which the robots' objective is to explore and collect reward in the environment. 
Some of the demonstrations in this application are found in~\cite{kolling2013human,howard2006experiments,kim2021plgrim,kaufmann2021copilot}.
Notably, our work has relevance on the MDP-based task allocation for Mars exploration with a rover and a copter~\cite{nilsson2018toward}.
While there exist different approaches in this domain, the study of multi-robot task allocation on real-world scenarios outside of the lab setup is still challenging due to uncertainty and complexity in environmental perception and robot capability modeling, and limited access to reliable communication means.

\ph{DARPA SubT Challenge} 
We choose to study our model in the context of the DARPA SubT Challenge as the competition requires the robot team to explore various extreme terrains and search for objects of interest in the environment. 
There were multiple teams participated in the challenge with different robot formations and autonomy solutions~\cite{agha2021nebula, cerberusJFR, csiroJFR, marbleJFR, explorerJFR, rouvcek2021system}.
However, the design of the robot formation and the operation strategy has been less of focus in literature due to the more imminent challenge in modular autonomy modules, such as communication~\cite{ginting2021chord, ginting2020deployable}, SLAM~\cite{ebadi2020lamp,chang2022lamp,blank2021autonomous}, global planning~\cite{kim2021plgrim, ginting2023safe}, and traversability analysis~\cite{fan2021step}.
In this work, we model the operation formally to infer optimal deployment strategy and team formation in a principled way.

\section{Problem Formulation}
\label{sec:problemsetup}
\fadhil{In this section, we model the environment, robot, communication constraint, and multi-robot task allocation problem.}
\subsection{Environment Model}

\begin{figure}
    \centering
    \includegraphics[width=0.4\textwidth]{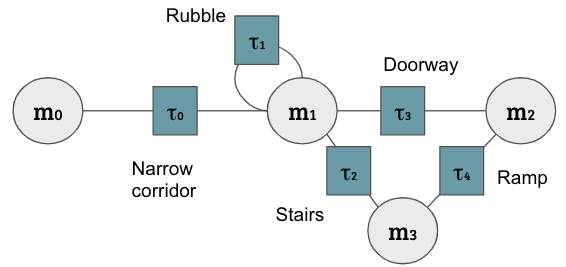}
    \caption{Topological map representation with traversability factors.}
    \label{fig:map_representation}
\end{figure}

\ph{Map structure} We represent the map of the environment as a graph denoted by $G = (V_{m}, V_{\tau}, E)$ as shown in \autoref{fig:map_representation}. $V_{m} = \{m_1, m_2, \ldots\}$ represents a finite number of segmented sectors in the environment, $V_{\tau} = \{\tau_1, \tau_2, \ldots\}$ represents the traversability hazards in the environment, and the edges $E$ define the connectivity between the sectors and traversability hazards.

\ph{Map belief} We assume that the map is not known \textit{a priori}. The robots gather the information about the sectors, traversability hazards, and their connectivity as they explore. Visiting each sector $m_i$ reveals information about the reachability of adjacent sector(s) if any, with the traversability requirements needed for the transition. In addition, the mission outcome associated with each sector is obtained based on the level of the information coverage within the sector.

\ph{Traversability representation}
We model the traversability from one sector to another as a stochastic process:
\begin{align}
    p(t_{i,j} \mid  \tau_{k}, \xi)
    \label{eq:trav_model}
\end{align}
where $t_{i,j}$ is the success transition probability from sector $m_i$ to $m_j$. 
We define $\tau_{k}$ as the specification of the traversability factor connecting two sectors and $\xi$ as the capability of the robot of interest. 

\ph{Reward representation}
The mission outcome is associated with each sector, and depends on the history of robot exploration behavior within the sector. The total mission reward $r_i$ from the sector $m_i$ is modeled as a distribution:
\begin{align}
    p(r_{i} \mid  m_{i}, c_{i})
\end{align}
where $c_i \in [0, 1]$ is the accumulated coverage level of the sector $m_i$. 



\subsection{Robot Model}

\ph{Robot state}
Let $x_i$ denotes the state of $i$-th robot in the team. A robot state is composed of geometric state $x^m$ (where the robot is), the health state $x^h$ (what is the health of sensing/mobility systems), and the mission state $x^r$ (how much unreported information the robot has).
\begin{align}
    x_i = \{x_i^{m}, x_i^{h}, x_i^{r}\}
\end{align}
We assume these robot states are fully observable. We also make an assumption that the health state is irreversible, meaning degraded sensors or mobility systems cannot be recovered during the mission, which is a reasonable assumption for exploring a non-human-accessible environment.

\ph{Robot capability}
The capability of the $i$-th robot is denoted by $\xi_i$ and characterized prior to the mission. $\xi_i$ is defined as 
\begin{align}
    \xi_i = \{\xi_i^{\tau}, \xi_i^{r}\}.
\end{align}
The mobility capability $\xi^\tau$ (e.g., legged, flying, and wheeled) is a dominating factor determining the exploration performance in a certain environment. Together with the traversability hazards specification, these factors determine the success probability of the traversal. The mission perception capability $\xi^r$ captures the ability to achieve the mission-critical perception. The mission perception capability might be limited for small payload systems, or might be degraded during the mission due to sensor damage.

\ph{Coverage model}
Robots with the mission perception capability can contribute to increase the coverage level within the sector defined as
\begin{align}
    c_{i}' = \sum_{j, \forall x^m_{j} = m_{i}} \Gamma(c_{i}, x_{j}),
\end{align}
where $c_{i},  c_{i}'\in [0, 1]$ is the coverage level of $i$-th segment at different time steps, $\Gamma(\cdot)$ is an update model given the previous coverage state and the current robot states.


\subsection{Communication Model}
In multi-robot exploration scenarios where there is no or limited communication infrastructure available, the robot team tends to require maintaining an ad-hoc networking system on the fly \cite{ginting2021chord}. We included such communication constraints in our process modeling.
In the typical mission setting, the robots need to report the mission-critical information back to the centralized base station through a communication link. 
We assume the robots and the base station can communicate within a certain range and line of sight using wireless devices. 
We represent sectors that are within the communication range to the base station as $V_o \subseteq V_m$. 
When the robots gather reward in sectors outside of the communication range, the robots need to return to $V_o$ to report the gathered reward.

\subsection{Multi-robot Task Allocation}

\ph{MDP formulation} We model the task allocation problem as an MDP represented as a tuple
\begin{align}
    (S, A, \mathcal{T}, \mathcal{R}),
\end{align}
where $S$ is a state space, $A$ is an action space, $\mathcal{T}(s'\mid  s, a)$ is a transition probability, and $\mathcal{R}(s)$ is a reward function.

\ph{State space} The state space consists of three factors: mapping $S^{m}$, robot status $S^{x}$ and mission progression $S^{r}$:
\begin{align}
    S = S^{m} \times S^{x} \times S^{r}
\end{align}
where $A \times B$ denotes a Cartesian product of two spaces $A$ and $B$; i.e., a set of all points $(a, b)$ where $a \in A$ and $b \in B$. The map state $s^m \in S^m$ holds the map belief on sector connectivity, traversability, coverage level, and communicability to the base station. The robot state $s^x \in S^x$ consists of the robot state $x_i$ for all fielded robots. The mission state $s^r \in S^r$ captures the mission progression such as accumulated rewards from the robot team.

\ph{Action space} 
Action space $A$ defines a set of actions that each robot can take. This is defined based on the mission and the robots' abilities. For example, our multi-robot exploration mission has the following action definition:
\begin{itemize}
\item $\texttt{frontier-seeking}$ directs the robot to move to a different sector at every time step according to a coverage policy.
\item $\texttt{guided-exploration}$ moves the robot to a certain area based on the commands from the human supervisor at the base station. It typically biases toward most-viable (rewarding) areas.
\item $\texttt{local-search}$ directs the robot to stay in the same sector and actively search in the environment to increase the coverage.
\item $\texttt{stay}$ keeps the robot in the same place, mostly to establish communications to other robots via the robot.
\end{itemize}




\ph{Transition probability}
The state transition model $\mathcal{T}(\cdot)$ can be decomposed by leveraging independency in the state space components and between different robots.
\begin{align}
    \mathcal{T}(s'\mid s, a) = Pr({s^m}'\mid  s, a) \left[ \prod_i Pr({s_i^x}'\mid  s_i, a_i) \right] Pr({s^r}'\mid  s, a)
\end{align}
where $i$ denotes the robot index. The map state transition models the update of our world knowledge based on the action taken by each robot. The robot state transition updates the robot position and health. The mission state transition updates the mission progression based on the updated coverage level.



\ph{Reward}
The reward model $\mathcal{R}(\cdot)$ is defined to encourage the robots to increase the coverage level $c_i$ in potentially rewarding sectors. A positive reward is given if the coverage level increases as the result of the robot's actions and no mission reward is previously awarded from the target sector.







\section{Policy Computation}
\label{sec:planning}
\subsection{Planning Objective and Constraints}
\ph{Objective}
The main objective of the mission is to gather as many rewards as possible in the environment until the termination condition is met. The reward $\mathcal{R}(s)$ can be gathered from the finite number of rewards available in the environment and can only be gathered by operational robots. The optimal policy of the deployment strategy is defined as a policy that maximizes the expected utility defined as 
\begin{align}
    \pi^*(s) = \argmax_{\pi} U^\pi(s).
\end{align}
The termination criteria of the operation are the loss of all operational robots (e.g., sensor/mobility failures), mission timeout, or full completion of the exploration.

\ph{Action constraints} 
When computing the policy, the action space $A$ is constrained by the capacity of the human supervisor or resources at the base station. We assume the human supervisor can only supervise one robot at each time step with semi-autonomous commands (e.g., \texttt{guided-exploration}). Each of the robot in the team can independently execute other fully-autonomous actions.

\begin{figure*}[t!]
  \centering
  \includegraphics[width=0.99\textwidth]{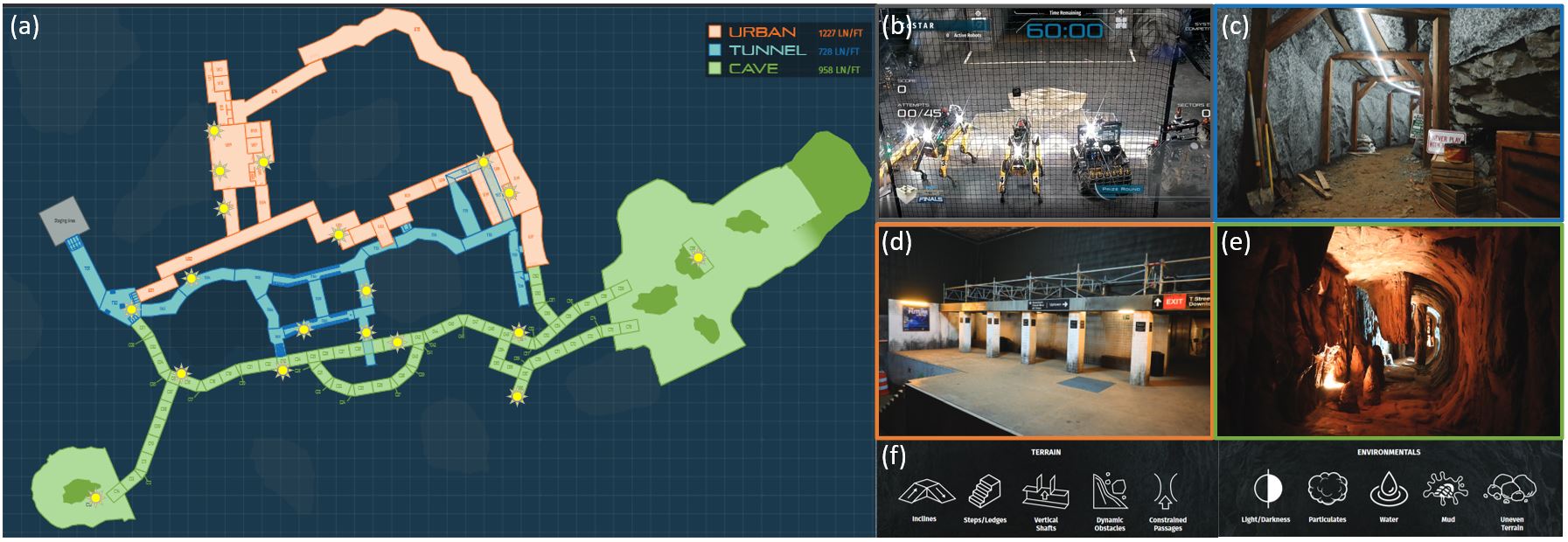}
  \caption{\textbf{(a)} The course map of the SubT Challenge Final Event which consists tunnel, urban, and cave environment. The location of the \textit{artifacts} are annotated in yellow. \textbf{(b)} Team CoSTAR's robots on the staging area (located on the top left of the map). \textbf{(c-e)} Photos from different part of the course: tunnel (c), urban (d), and cave (e) environment. \textbf{(f)} Different type of terrain and environmental challenges in the course. The base map and photos are provided by DARPA. The annotations are added by the authors. }
  \label{fig:final_event_setup}
\end{figure*}

\subsection{Receding Horizon Planning}
\fadhil{
We use the receding horizon planning approach for two reasons. 
First, the map state $S^m$ is not known in advance and grows as the robot team explore the environment, so we can only plan for a limited horizon. 
Second, the state and action space for planning multiple robots in this problem is large and online planning reduces the computational requirements for the planner to be useful in real-world operations. 
In receding horizon planning, we plan from the current state to some fixed horizon. Then, we execute the best policy and replan.

We use the Monte Carlo Tree Search (MCTS) approach that runs simulations from the current state and builds a search tree. 
Particularly, we use Double Progressive Widening (DPW) method that enable the simulation to search deeper in the tree for MDP with large state and action spaces~\cite{couetoux2011dpw}. 
In this work we use \texttt{MCTS.jl} package in Julia and define our MDP with \texttt{POMDP.jl}~\cite{egorov2017pomdps}. 
}

\section{Simulation Results and Case Study}
\label{sec:simulation}
The developed multi-robot mission model is evaluated on the DARPA SubT Challenge scenario. 
We evaluate the computed deployment policy from our model and compare it against the policy developed by the human team in the live real-world operation. 
Moreover, we analyze the impact of the robot's number and autonomy capability to the mission performance of the robot team.






\subsection{Modeling Heterogeneous Multi-robot Deployment}
\label{subsec:exp-model}

\ph{Scenario}
The DARPA SubT Challenge seeks novel approaches to rapidly map, navigate, and search underground complex environments.  
The participating teams are evaluated by the number of discovered \textit{artifacts} that are the main object of interests in the search and rescue application. The teams must send robots to explore, search, and localize the artifacts in the km-scale environments. 

\ph{Concept of operation}
In this work, we model NASA JPL-led Team CoSTAR's field operation in the SubT Challenge.  In the operation, Team CoSTAR deploys a team of autonomous robots to explore and map the large-scale environment \cite{kim2021plgrim,ebadi2020lamp}. At the same time, the robots build communication mesh networks to transmit detected artifact information to the base station \cite{ginting2021chord}. Team CoSTAR deploys a team of robots of heterogeneous capabilities. For example, the team deploys legged robots to traverse stairs and drones to enter vertical shafts. For more information about Team CoSTAR's robotics and autonomy solution, see \cite{agha2021nebula}.

\ph{Environment model} 
We build the environmental model based on the Final course specification provided by DARPA~\cite{subt_final}.
As illustrated in \autoref{fig:final_event_setup}, the course is divided into sectors that have certain terrain and environmental characteristics.
There are a finite number of artifacts distributed in the course.
For example, in the Final event course, there are 40 artifacts scattered in 121 sectors.
In every sector, we define two level of coverage whether the sector has been searched intensively to find an artifact hidden in the section.


\ph{Heterogeneous robot model}
We model three different robot types that Team CoSTAR deployed in the competition: wheeled robot, legged robot, and flying robot (See \autoref{fig:robot_team_in_mine}). Each robot has the same action spaces but is associated with different state transition probabilities. The robot transition probability to move between sectors depends on the terrain type in the course and the mobility capability of the robot, as modeled in \eqref{eq:trav_model}.
If the robot fails to transit between the sectors, the robot will break and will not be operational for the rest of the mission.

\ph{Reward model}
The robot team receives 1 reward point for every artifact the robot discovered in the environment. The mission state $s^r$ tracks the number of accumulated reward reported to the base station.


\subsection{Case Study: Urban Building}
\label{subsec:exp-urban}

\begin{figure}[t!]
  \centering
  \includegraphics[width=0.45\textwidth]{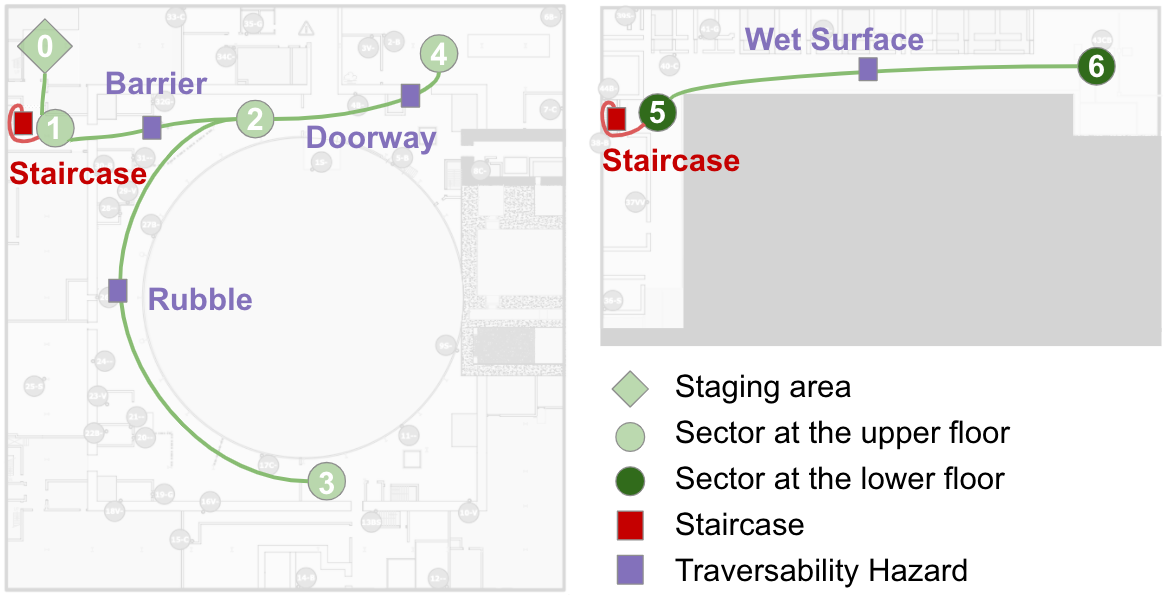}
  \caption{Urban building exploration scenario inspired by DARPA SubT Challenge Urban Circuit. 7 sectors in two floors are connected with a staircase. Two robots (wheeled, legged) are deployed from the staging area. }
  \label{fig:urban_map}
\end{figure}



\ph{Scenario} 
To demonstrate that our model can compute a deployment policy for a robot team with heterogeneous capabilities, 
we consider a simple scenario where two robots explore a two-floor building as shown in \autoref{fig:urban_map}. 
The robot team consists of one legged robot and one wheeled robot. 
The environment model was generated based on the power plant map used in the Urban Circuit of the SubT Challenge.
The two-floor building is divided into 7 sectors ($|V_m| = 7$) and a staircase connects the sectors in the different floors. 
In this scenario, there is a high probability a legged robot can climb the stair successfully while the wheeled robot has a high chance of failures. 
The robots need to search for an artifact in each sector and report the detected artifacts to the base station. 
The wireless communication coverage is limited to nearby sectors to the base station $V_o = \{m_i \mid  i=0 \ldots 5\}$ and the robots need to return to sectors with communication coverage to report the artifacts and earn the mission rewards. 
The planning horizon in this experiment is 5 time steps.

\begin{figure}[t!]
  \centering
  \includegraphics[width=0.49\textwidth]{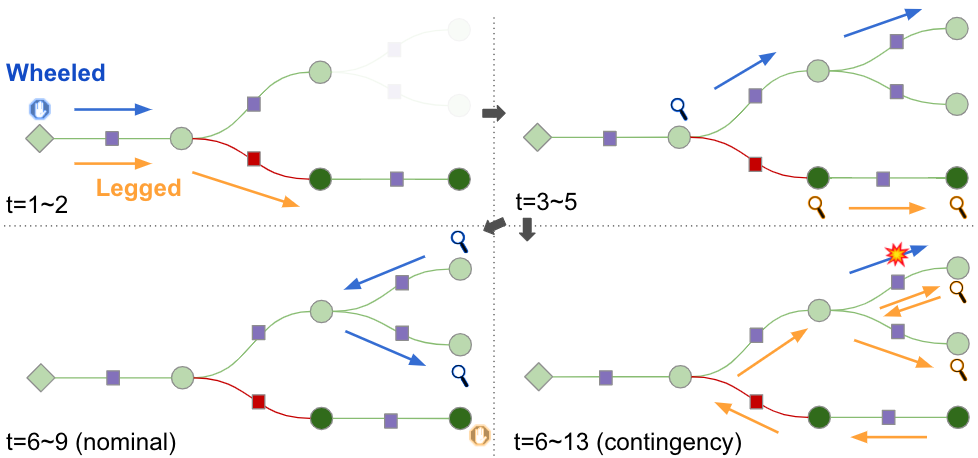}
  \caption{Visualization of robot team behaviors in nominal and contingency cases for the urban building scenario. In the nominal case, the optimal policy directs the legged robot to explore the lower level through stairs while the wheeled robot to explore the upper level. If the wheeled robot fails to move between sectors due to a traversability hazard, the legged robot on the lower level is guided to explore the remaining upper-floor sectors.}
  \label{fig:urban_plan}
\end{figure}

\ph{Evaluation of optimal policy} 
\autoref{fig:urban_plan} shows the optimal deployment using the computed policy. At a high level, the optimal policy sends the legged robot with the stair climbing capability to explore the lower level while having the wheeled robot gather rewards on the upper level. 
The strategy matches with the Team CoSTAR's deployment in Urban Circuit \cite{bouman2020spot} where the team won the first place by scoring artifacts from multiple floors. The policy also suggests when the human supervisor should supervise the robots to increase the exploration performance: for example, the supervision is effective when the robots are exploring nearby the lethal traversability hazards (e.g., stairs for wheeled robots) to maintain safety. In addition, the offline policy provides a contingency plan in case one of the robot is damaged. For example, if the wheeled robot fails, the policy guides the legged robot to the upper floor to continue collecting the remaining mission reward.

\ph{Comparison against naive policies}
To evaluate the significance of the optimal policy, we compare the mission performance of our approach against naive exploration policies. The statistics are summarized in \autoref{table:1}. The \textit{full supervision} policy is computed based on a limited action set that only allows the robots to move based on the supervisor's supervision commands.
\begin{align}
A_{\textrm{supervision}} =
\begin{Bmatrix}
    \texttt{guided-exploration}\\
    \texttt{stay}\\
    \texttt{local-search}
\end{Bmatrix}
\end{align}
The \textit{naive autonomy} policy lets the robot exploring the environment by itself without supervision.
\begin{align}
A_{\textrm{naive}} =
\begin{Bmatrix}
    \texttt{frontier-seeking}\\
    \texttt{stay}\\
    \texttt{local-search}
\end{Bmatrix}
\end{align}
The \textit{random} policy picks a random action for every time step among the all available actions. 

Our approach generates a deployment policy with higher expected rewards, and a lower number of robot failures compared to the naive autonomy policy and random policy. While the full supervision provides the same expected reward and the number of robot failures as ours, it takes twice as long to complete the exploration since each robot needs to be manually commanded all the time.

\begin{table}[t]
\caption{Comparison between our supervised autonomy approach and naive policies in different mission performance metrics. The maximum number of expected rewards in this case study is 4. The maximum number of expected robot failures is 2. The expected value are estimated from 1000 simulations.}
\centering
\begin{tabular}{lrrr} 
 \toprule
  & Rewards & Robot Failures & Mission Time\\
 \midrule
 \textbf{Our Approach}  &   \textbf{3.58}  &  \textbf{0.59} &  \textbf{8.15}  \\
 Full Supervision &   \textbf{3.58}  &  \textbf{0.59} &  16.57  \\
 Naive Autonomy  &   2.46  &  1.68 &  24.23  \\
 Random    &   1.04  &  1.97 &  33.71  \\ 
 \bottomrule
\end{tabular}
\label{table:1}
\end{table}
\subsection{Case Study: SubT Challenge Final Event}
\label{subsec:exp-real-world-study}

\ph{Scenario}
To show that our operation model can be applicable to complex real-world scenarios, we evaluate it on our field operation on the Final event of the SubT Challenge (\autoref{fig:final_event_setup}). 
The course consists of 121 sectors that possesses Tunnel, Urban, and Cave features, and spans around 900\,m in length.
The DARPA Final Course Callouts describes the terrains, environmental components, and the difficulty rating on every sector~\cite{subt_final}.
In the live operation, we deployed 7 robots consisting of 4 legged robots, 2 wheeled robots, and 1 flying robot. 


\ph{Simulation results}
We first evaluate the policy performance based on the simulated episodes.
With the computed policy, the robot team is able to explore $73 \%$ of the sectors on average, finding 14 artifacts.
The policy directs the legged robots to the Urban with narrow passages and stairs and bumpy Cave area and while wheeled robots are sent to the Tunnel area, reflecting the terrain preference of the locomotion systems.
Interestingly, it does not send all the robot at the same time because of the limited supervision capacity by the human supervisor. 
This margin allows the human supervisor to intervene potential unsafe robot behaviors at difficult sectors. 
We observe different exploration behaviors for robots depending on the terrain difficulties. For safer sectors, the policy directs the robots to go deep first to discover as much sectors/artifacts as possible before returning back to sectors within the communication range. 
Meanwhile, robots exploring difficult sector return to the communication range more frequently as the robots have higher risk of failures when exploring in these sectors.

\ph{Comparison with human team} 
We compare the computed operation policy with the live real-world operation by the human team during the SubT Final event in September 2021\footnote{The video of our real-world operation is available at \url{https://youtu.be/4CEQ814UhGY}}. 
On day 2, Team CoSTAR scored 11 artifacts out of 20 and explored $60 \%$ of the course which is close to the statistics in the simulation. 
The human deployment strategy in the competition was similar to the strategy derived from the developed operation model. 
The human team first deployed 3 legged robots with guided exploration as they are more capable of finding new sectors/artifacts by traversing challenging terrain components in the course. The wheel robots are kept as backups for the time when the legged robots fail or discover extra areas to explore.
From the team's experience in deploying autonomous heterogeneous robots into the field, our operation model can capture different factors that human team considers to plan for the robot deployment into unknown complex environments.

\fadhil{
\ph{Benefits of multi-robot task allocation in real-world operations} 
When applying our model to the real-world operations, there are two main benefits that our model provides to support human-team decision making. 
First, our model can compute task allocation policies that incorporate more information than the human team, considering the human's limited ability to process numerous data. 
To oversee multi robots, the human supervisor only monitor high-level information reported by robots such as robot's health, detected artifacts, and sparse 3D point cloud map.  
The human-based strategies are typically too simplified and limited (e.g., bind specific robot types to specific areas) because of the lack of understanding of the remote environments and the limitation in predicting various future possibilities. 
On the other hand, our model can compute a more detailed deployment policy by considering terrain variations within each environment. 
Second, our model can anticipate when a robot needs supervision by the human supervisor. 
Due to the limited capacity of the human supervisor to oversee 7 robots, the human supervisor only intervenes a robot when there is a warning notification of potential unsafe behaviors. By computing the policy consisting of sequence of robot actions, our model can anticipate and inform the human supervisor in advance when a robot will traverse around unsafe areas. 
For the system implementation, the operation model can provide the task allocation policy and support the human team through a task scheduler and human-robot interface~\cite{kaufmannCopilot}.
}


\begin{table}[t]
\caption{Comparison of robot formation in the case study based on the mission performance metrics}
\centering
\begin{tabular}{lrrrr} 
 \toprule
 & \multicolumn{2}{c}{Robot Formation} &  \\
 \cmidrule{2-3}
 & Wheeled & Legged & Reward & Robot Failures \\
 \midrule
 Multi hybrid & 1 & 1 &   \textbf{3.58}  &  \textbf{0.59} \\ 
 Multi wheeled & 2 & 0 &   2.56  &  1.53 \\ 
 Multi legged & 0 & 2 &   \textbf{3.61}  &  0.60  \\ 
 Single wheeled & 1 & 0 &   2.07  &  0.86 \\ 
 Single legged & 0 & 1 &   2.57  &  \textbf{0.59} \\ 
 \bottomrule
\end{tabular}
\label{table:2}
\end{table}




\section{Team Formation Analysis}
\label{sec:exp-capability-analysis}
In this section, we show that our multi-robot operation model can be used to determine the best robot team configurations given the scenario description. 
Moreover, we discuss how the model can be used as a design tool to prioritize the development effort of the robot capabilities. 
We focus on four areas of robot capabilities: mobility, perception, communication, and autonomy.

\ph{Mobility configurations}
We compare the mission performance for different robot team configurations on the urban building scenario in \autoref{subsec:exp-urban}.
\autoref{table:2} summarizes the expected rewards and robot failures for different robot team configurations.
The result indicates that higher number of robots is better in terms of accumulating rewards under the risk of immobilization due to traversability hazards. Despite the high counts of robot failures, more robots provide the system resiliency to a single failure event.
However, it should be noted that while deploying more robots is beneficial, deploying more than what the scale of environments requires will not add much improvement to the mission performance. Rather, in the real-world applications, it might pose coordination challenges such as congestion control, as Team CoSTAR experienced on the first day of the Final competition.
Diversification of the mobility systems helps to improve the robustness to various types of mobility hazards. Different sectors in the environment have mobility challenges and particular types of robots are suitable to traverse specific terrains. By forming a team of heterogeneous mobility systems, one can avoid hitting the exploration bottleneck caused by certain types of environmental challenges. If the characteristics of the target environment is known a priori, forming a team of specialized mobility system can increase the potential reward obtained at the cost of being less robust to different terrain configurations.

\ph{Perception capability}
Perceptual challenges are one of the main causes that significantly hinder the mission execution, particularly in perceptually-degraded environments such as subterranean structures. Examples from our field experience include sudden lighting changes, particulates, recovering from sudden movement (e.g., fall), and self-similar structures. The fatal perception failures often lead to the loss of the robots, hence significantly affecting the mission performance.
Our mission modeling captures the perception failures as probabilistic events whose occurrence is a function of robot's capability. The expected mission reward increases as the failure probability decreases, until the resiliency of having multiple robots surpasses the need for a single robot to be fully self-contained.
There is an interesting system trade-off about how to achieve the resiliency, that also depends on the cost of development and operations. The study of such trade-off is left for future work.

\begin{figure}[t!]
  \centering
  \includegraphics[width=0.40\textwidth]{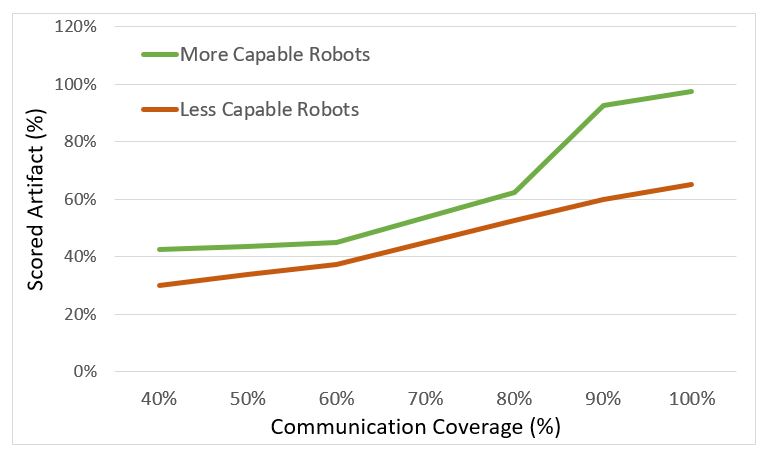}
  \caption{Trade-off between communication coverage and robot capability on the mission performance. The communication coverage is the percentage of the sectors in the course within the communication range. The mission objective is the percentage of scored artifacts.}
  \label{fig:comm_autonomy_tradeoff}
\end{figure}

\ph{Communication vs autonomy trade-off}
The robots outside of the communication range have a huge risk of mission termination without communicating back the collected mission reward to the base station. There are two major approaches toward this problem: extending the communication coverage so that the robots are always kept in the communication range, or improving the robot's autonomy capability so that they can reliably operate for an extended period of time. 
We performed an analysis that looks into the relation between the communication coverage and robot autonomy capability.
\fadhil{\autoref{fig:comm_autonomy_tradeoff} compares the mission performance for different communication coverage and robot capabilities on the SubT Challenge Final Event scenario in \autoref{subsec:exp-real-world-study}}. 
\fadhil{First of all, having a good communication coverage is essential to achieve the high mission performance. Even with the best autonomy, the robots will still be lost due to the mobility failures, perception challenges, hardware issues, and so on. Maintaining good communication to the robots and frequently reporting the discovered mission reward is important to avoid information loss. 
Secondly, at the high communication coverage, having capable robots makes significant difference on the mission performance as the high autonomy capability allows the robot team to explore larger areas. The increase in the communication coverage does not significantly contribute to the mission performance once it exceeds the certain level (around 90\% in this analysis) due to the high returning rate from the non-communicable area. The communication and autonomy play a complementary role in terms of achieving the better mission performance, and the development effort needs to be balanced between two.}

\section{Conclusions}





This paper studied task allocation and team formation problems for multi-robot cooperative exploration in complex previously-unknown environments. 
We developed a MDP-based model to describe the missions and to make deployment and commanding decisions based on the latest map, robot, and mission states. 
\fadhil{We performed a case study using the DARPA SubT Challenge deployment scenarios and showed our operation model generates comparable deployment strategies with the human-based strategy performed by Team CoSTAR during the competition. 
Moreover, the operation model can compute a more detailed strategy and support human-team decision making in real-world operations. }
\fadhil{The developed mission model, accompanied by team CoSTAR's three-year of experience in deploying multi-robot system, 
is used to inform the design of a team of robots to perform an exploration task in real-world with resilience and robustness.}

\renewcommand*{\bibfont}{\small}
\printbibliography

\end{document}